\documentclass[preprint]{article}

\usepackage{microtype}
\usepackage{graphicx}
\usepackage{subcaption}
\usepackage{booktabs} 

\usepackage{hyperref}

\usepackage{icml2026}

\usepackage{amsmath}
\usepackage{amssymb}
\usepackage{mathtools}
\usepackage{amsthm}
\usepackage{enumitem}

\usepackage[capitalize,noabbrev]{cleveref}

\theoremstyle{plain}

\theoremstyle{definition}

\theoremstyle{remark}

\usepackage[textsize=tiny]{todonotes}

\icmltitlerunning{Context is All You Need}

\begin{document}

\twocolumn[
  \icmltitle{Context is All You Need}



  \icmlsetsymbol{equal}{*}

  \begin{icmlauthorlist}
  \icmlauthor{Jean Erik Delanois}{ucsdcs}
  \icmlauthor{Shruti Joshi}{ucsdmed}
  \icmlauthor{Ryan Golden}{ucsdmed}
  \icmlauthor{Teresa Nick}{msoft}
  \icmlauthor{Maxim Bazhenov}{ucsdmed}
  \end{icmlauthorlist}

\icmlaffiliation{ucsdcs}{Department of Computer Science \& Engineering, University of California, San Diego, La Jolla, California, USA}
\icmlaffiliation{ucsdmed}{Department of Medicine, University of California, San Diego, La Jolla, California, USA}
\icmlaffiliation{msoft}{Microsoft Corporation, Redmond, WA, USA}

\icmlcorrespondingauthor{Maxim Bazhenov}{mbazhenov@ucsd.edu}

  \icmlkeywords{Machine Learning, ICML}

  \vskip 0.3in
]



\printAffiliationsAndNotice{}  

\begin{abstract}
 Artificial Neural Networks (ANNs) are increasingly deployed across diverse real-world settings, where they must operate under data distributions that differ from those seen during training.  This challenge is central to Domain Generalization (DG), which trains models to generalize to unseen domains without target data, and Test-Time Adaptation (TTA), which improves robustness by adapting to unlabeled test data at deployment. Existing approaches to address these challenges are often complex, resource-intensive, and difficult to scale.
 We introduce \textsc{CONTXT} (\emph{\textbf{C}ontextual augmentati\textbf{O}n for \textbf{N}eural fea\textbf{T}ure \textbf{X} \textbf{T}ransforms}), a simple and intuitive method for contextual adaptation. \textsc{CONTXT} modulates internal representations using simple additive and multiplicative feature transforms. Within a TTA setting, it yields consistent gains across discriminative tasks (e.g., ANN/CNN classification) and generative models (e.g., LLMs). The method is lightweight, easy to integrate, and incurs minimal overhead, enabling robust performance under domain shift without added complexity. More broadly, \textsc{CONTXT} provides a compact way to steer information flow and neural processing without retraining.
\end{abstract}

\section{Introduction}


Artificial neural networks (ANNs) today power image, speech, recommendation, and text systems, but as they scale into products their failure modes become increasingly evident. A key practical challenge is domain shift: models trained in one context often lose performance in another, encompassing distribution shift, out-of-distribution (OOD) behavior, spurious correlations, and context misalignment. In practice, practitioners frequently need a classifier to adapt to a new deployment domain or a generator to produce context-appropriate outputs under changing conditions. The root cause is a train–deploy mismatch: models optimize for the training context and then encounter a different one at test time. 
Large language models (LLMs) show a similar brittleness: they can over-rely on training priors, prompt phrasing, system instructions, or retrieved passages, and therefore often underperform when task conditions shift unless the context is appropriately updated. 

At a broader level, this highlights a core limitation of current ANNs: incorporating new knowledge or adapting to novel contexts typically requires complex contextual architectures, fine-tuning, or full retraining on new data. Fine-tuning risks catastrophic forgetting \citep{french1999catastrophic,hayes2021replay,luo2024}, and retraining large models is costly and inefficient. 
Since 2012, state-of-the-art training compute requirements have doubled about every 3.4 months (roughly $10\times$ per year), outpacing Moore’s law \citep{openai2018compute,sevilla2022compute} and driving an unsustainable long-term increase in energy and water use. 

\paragraph{Examples of DA/TTA and contextual sensitivity:} In vision, classic domain-shift benchmarks reveal brittleness across style, texture, and environment, including PACS (Photo, Art, Cartoon, Sketch) \citep{li2017deeper}, Office-Home \citep{venkateswara2017deep}, Terra Incognita \citep{beery2018recognition}, and the WILDS benchmark suite \citep{koh2021wilds}. For LLMs, small changes in instructions or retrieved context can alter output style, safety posture, and depth of reasoning. Carefully crafted prompts can elicit toxic or hateful speech, as shown in HarmBench \citep{mazeika2024harmbench}. Under adversarial context, LLMs can be jailbroken to perform undesirable tasks, including behaviors they were explicitly trained to avoid \citep{chao2024jailbreakbench}.

\paragraph{Existing approaches and their practical limitations:} A large literature addresses domain shift using multiple strategies. Representative families include: (i) data-centric augmentation and style randomization (AugMix, RandAugment, Stylized-ImageNet) \citep{hendrycks2020augmix,cubuk2020randaugment,geirhos2019imagenet}; (ii) representation alignment and invariance penalties (Deep CORAL, MMD-based methods, domain-adversarial training) \citep{sun2016deepcoral,li2018mmd,ganin2016dann}; and (iii) objective- and test-time adaptation methods targeting worst-case or online shifts (GroupDRO/REx, TENT, test-time BN) \citep{sagawa2020distributionally,krueger2021rex,wang2021tent,schneider2020ttbn}. Interpretable-prototype architectures have also been proposed for robust class-specific representations \citep{chen_this_2019}, but they require retraining and are limited to supervised image classification. While effective, many of these approaches rely on extensive engineering, auxiliary models, or fragile test-time optimization, limiting deployment in resource- or latency-constrained settings.

Activation-engineering and steering methods offer a lower-cost alternative by directly manipulating internal activations to bias outputs \citep{turner2023steering,cheng2024linearly,panickssery2023steering}. However, they typically rely on token-level offsets or paired prompts, assuming clean conceptual opposites and precise token alignment. This makes them brittle for abstract concepts, sensitive to tokenization or length mismatches, and prone to fading effects in long generations. Other methods inject lightweight biases \citep{subramani2022extracting}, but still require backpropagation-based training prior to inference. 


\vspace{-6pt}
\paragraph{The brain as a guide to context.} 
 Biological systems routinely handle shifts in context. Humans recognize a chair whether it is photographed, sketched, or described verbally; we adapt to new lighting or furniture, and switch conversational registers from technical to casual without explicit retraining. While we still lack a complete understanding of how the brain achieves this, neuroscience data points to complex bi-directional interactions between the prefrontal cortex (PFC), hippocampus, and sensory areas.
 
 One influential idea was proposed by Nadel and Willner, who introduced a Dual-Process Theory of context representation \citep{NadelWillner1980,Rudy_2009}. In this framework, the prefrontal cortex supports relatively inflexible, feature-based conditioning, while the hippocampus and associated structures form a relational system specialized for encoding and recognizing context \citep{PENG2021107520}.
 

In line with this framework, the PFC is critical for retrieving context-relevant memories \citep{Navawongse1002} and suppressing context-irrelevant sensory inputs \citep{Farovik13428}. Complementary work shows that the hippocampus rapidly encodes conjunctive representations of context \citep{Yadav_Noble_Niemeyer_Terceros_Victor_Liston_Rajasethupathy_2022} and tracks contextual consistency across space and time, including detecting boundaries between distinct contexts \citep{DAVACHI201592,Maurer_Nadel_2021}. These roles are tightly coordinated: contextual information flows from hippocampus to PFC upon entry into a new context, but reverses direction during subsequent environmental sampling \citep{Place_Farovik_Brockmann_Eichenbaum_2016}.

Together, this suggests that the hippocampus encodes and identifies the current context and relays this information to the PFC, which then exerts top-down control to reinterpret identical inputs as context changes, without requiring new learning \citep{HalassaKastner2017,Schmitt2017MD,Stokes2015DynamicCoding}. Accordingly, neurobiological context modulation can be viewed as two sequential processes: (1) rapid hippocampal encoding of context and (2) PFC-mediated amplification and suppression of contextually relevant and irrelevant features. In this work, we assume access to previously encoded contextual representations and focus on how a PFC-like module can implement the second process. Explicit modeling of hippocampal context discovery is left for future work and discussed in the conclusion.


\begin{figure*}[!ht]
  \centering
  \begin{subfigure}[t]{0.23\linewidth}
    \centering
    \includegraphics[width=\linewidth]{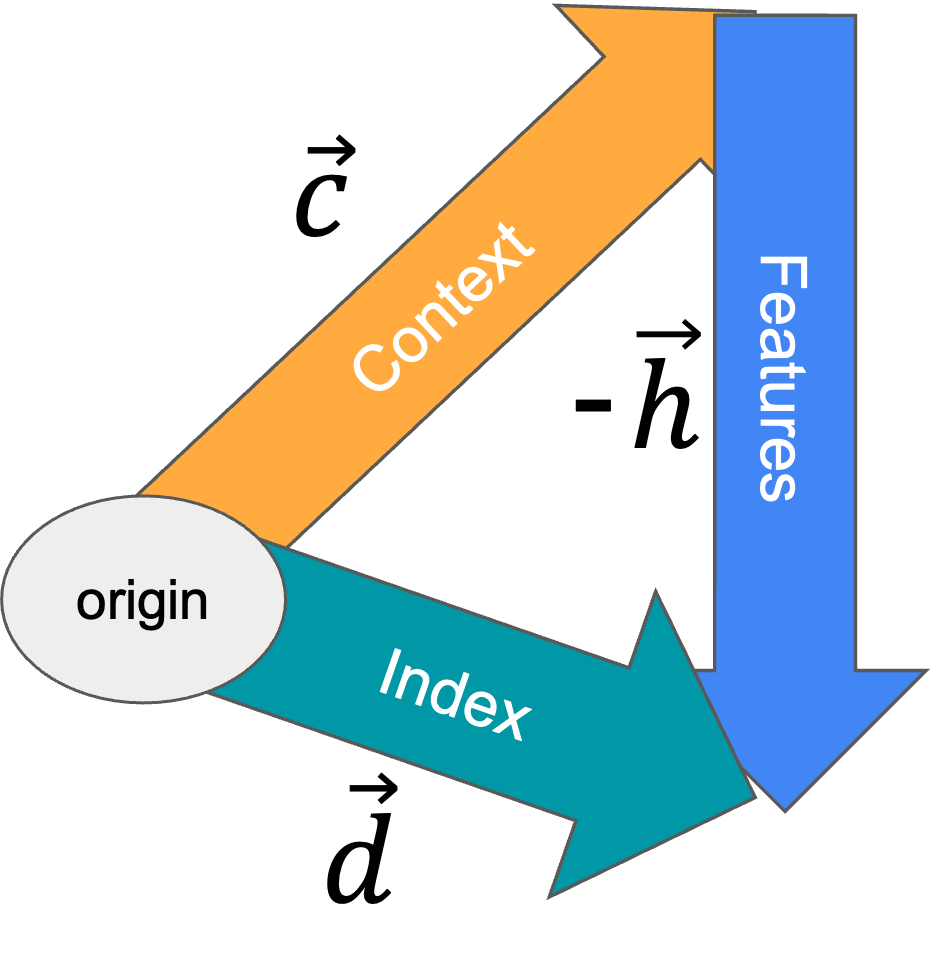}
    \caption{Index Computation}
    \label{fig:two-up:a}
  \end{subfigure}
  \begin{subfigure}[t]{0.2\linewidth}
    \centering
    \includegraphics[width=\linewidth]{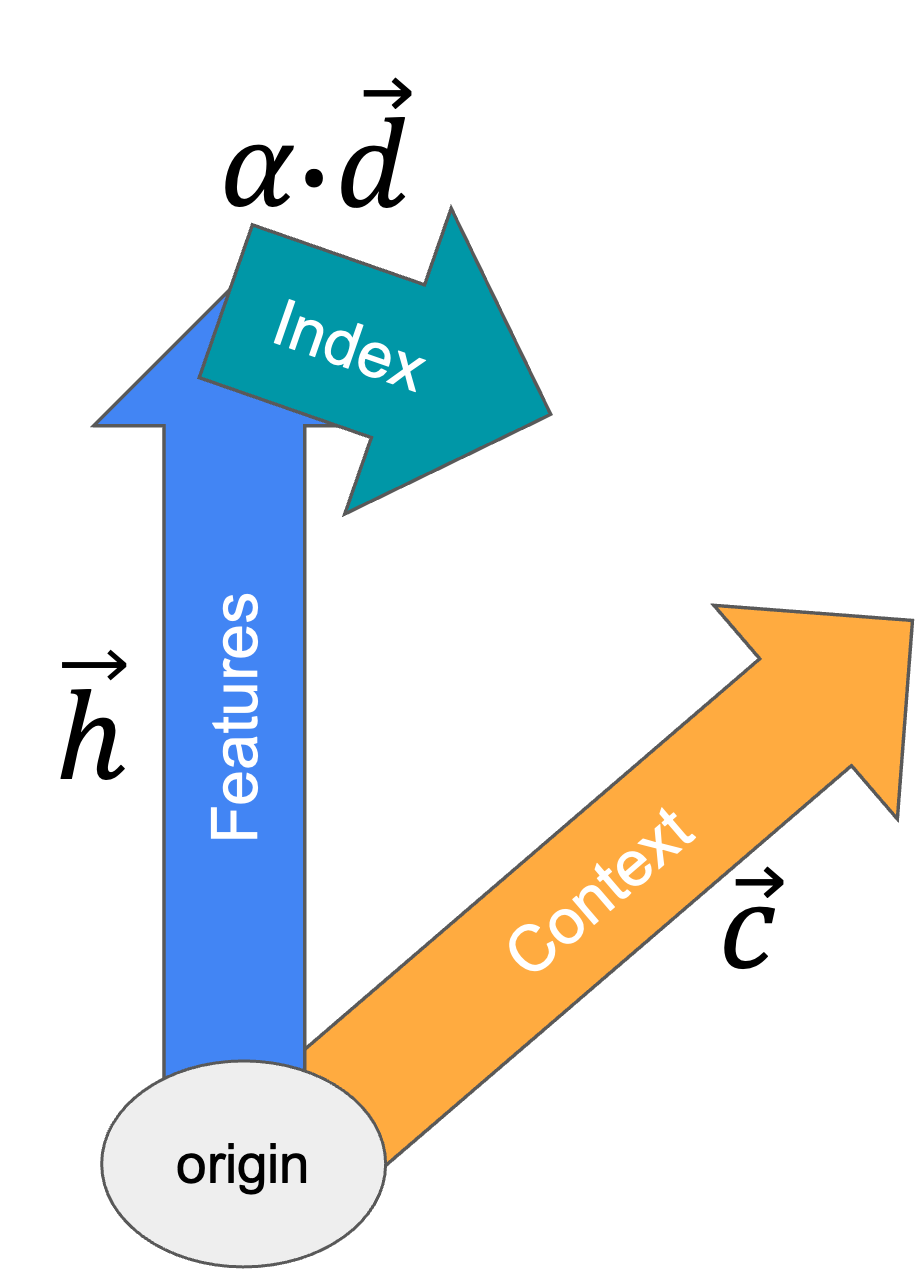}
    \caption{Index Application}
    \label{ }
  \end{subfigure}
  \begin{subfigure}[t]
  {0.3\linewidth}
    \centering
    \includegraphics[width=\linewidth]
    {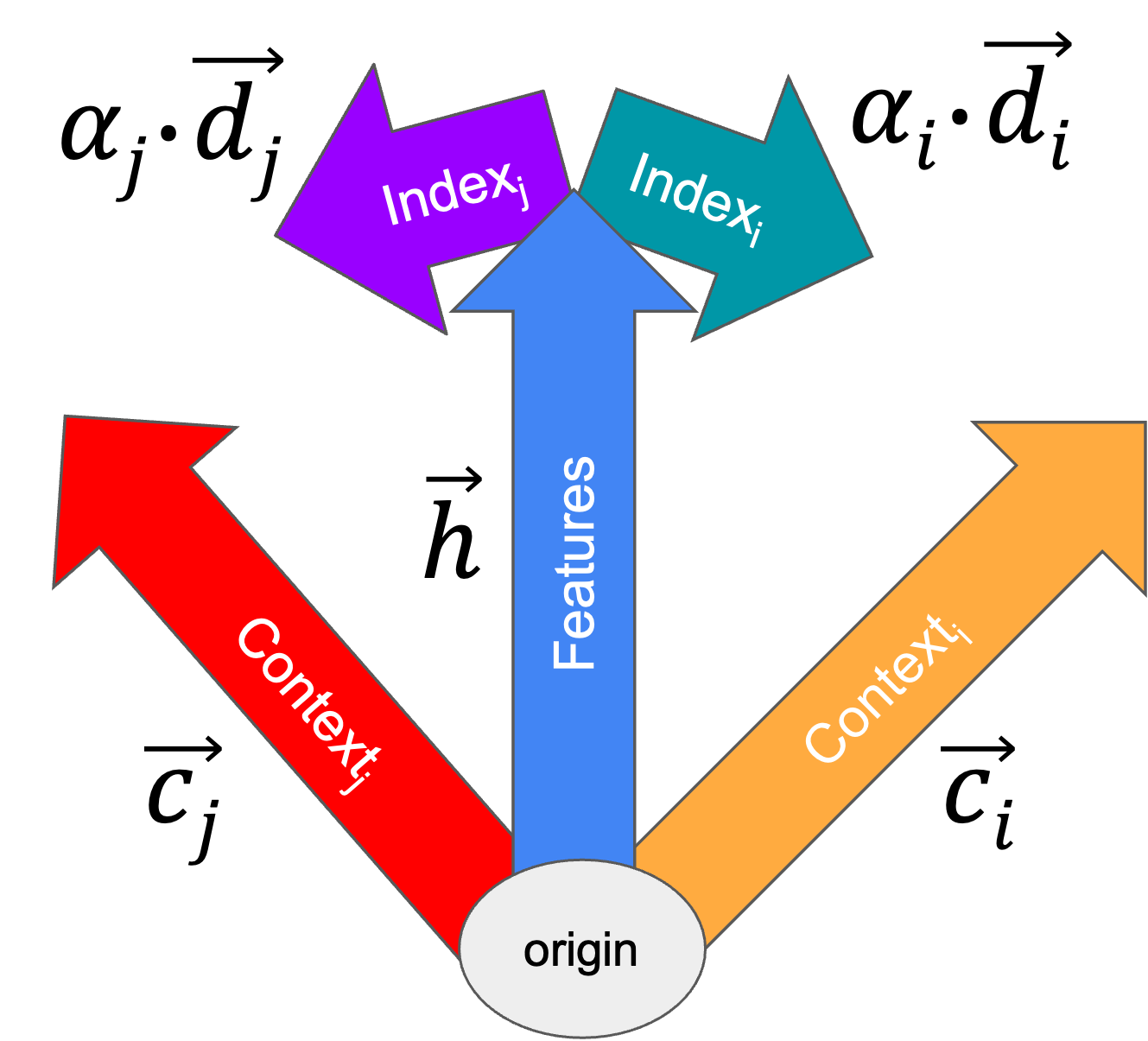}
    \caption{Multi Index Application}
    \label{ }
  \end{subfigure}
    \caption{\textbf{\textsc{CONTXT}: Contextual augmentation via feature transforms.} (a) At a chosen layer, compare the current feature vector $\mathbf{h}$ to a precomputed contextual feature representation $\mathbf{c}$ to form an "index" (their difference) $\mathbf{d}=\mathbf{c}-\mathbf{h}$. (b) Add a scaled version of this index, $\alpha\mathbf{d}$, to the features; $\alpha>0$ injects the context while $\alpha<0$ removes it. (c) Mix multiple contexts by linearly combining indices with separate scalars, e.g.\ $\alpha_i\mathbf{d}_i+\alpha_j\mathbf{d}_j$.}

  \label{fig:contxtDiagram}
\end{figure*}

\paragraph{Main contributions:} ANN feature spaces can exhibit strikingly linear, human-interpretable structure; famously, \emph{king} $-$ \emph{man} $+$ \emph{woman} $\approx$ \emph{queen} \citep{mikolov2013linguistic}. Despite the ubiquity of this intuition, it has been under-utilized for improving downstream performance. A handful of works leverage linear directions to steer generative models \citep{turner2023steering, cheng2024linearly, subramani2022extracting, panickssery2023steering}, but these approaches are often specialized or cumbersome, and comparable solutions for classification are largely absent. We introduce \textsc{CONTXT} (\emph{\textbf{C}ontextual augmentati\textbf{O}n for \textbf{N}eural fea\textbf{T}ure \textbf{X} \textbf{T}ransforms}), a simple, lightweight mechanism for contextual adaptation that applies across common layers and architectures. A \textsc{CONTXT} layer combines current feature representations with saved, context-specific features to form an index vector, which then augments the current features via simple multiplicative and additive operations, steering processing based on the active context.

Because \textsc{CONTXT} operates on internal representations rather than model weights, it is parameter- and compute-efficient and integrates easily into existing pretrained networks. In practice, it can improve classification by downweighting unfamiliar contextual cues and injecting familiar ones, and it can bias generative models toward context-appropriate outputs without retraining or prompt engineering. 
Our specific contribution is a mechanism for contextual adaptation during inference that:
\begin{itemize}[noitemsep]
    \item Requires no additional training, sparse autoencoders, or paired prompts,
    \item Operates directly in feature space via lightweight vector arithmetic, and
    \item Applies to both discriminative and generative models without architecture-specific modifications.
\end{itemize}

Compared with retraining or domain-specific fine-tuning, \textsc{CONTXT} requires only two forward passes (context and input) and lightweight vector arithmetic. Unlike activation-steering methods that rely on token-level alignment or backpropagation during generation \citep{Turner2023ActivationSteering,Zou2023RepEng,Dathathri2020PPLM}, it uses a individual contextual feature representations and a scalar weights applied across hidden layer activations and tokens. The method requires minimal engineering, scales across deployment settings, and can be toggled with negligible effort and cost, while still delivering substantial gains for both classification and generation tasks.



\section{Methods}\label{sec:Methods}

\textbf{Contextual augmentation for Neural feature X Transforms (\textsc{CONTXT})} modifies intermediate network features by injecting or removing contextual information, thereby altering model behavior without retraining. In classification, \textsc{CONTXT} can improve performance under domain shift (e.g., adapting a wildlife classifier trained in urban settings to the beach settings by shifting away from the “beach” context and toward the “urban” context). 
For LLMs, \textsc{CONTXT} can impart sentiment or high-level concepts without changing the prompt.

\paragraph{Operation.}
Let $h_\ell(x)\in\mathbb{R}^d$ denote the feature representation of input $x$ at layer $\ell$. For a context $\kappa$, we precompute a \emph{context vector} $c_{\ell,\kappa}$ at the same layer—either the feature of a representative sample (for LLMs) or the mean feature over samples exhibiting context $\kappa$  (for classification models). We assume here that we know the context or domain in which the model was trained as well as we know the context from which we get the testing data (but see Discussion for relaxing this assumption). Given $h_\ell(x)$ and $c_{\ell,\kappa}$, we form a \textsc{CONTXT} \emph{index}
\[
d_{\ell,\kappa}(x) \;=\; c_{\ell,\kappa} - h_\ell(x) \quad \text{(Figure \ref{fig:contxtDiagram}a)}.
\]
This index defines a direction in activation space along which the representation can be shifted to bring it closer to the training context or moved away to suppress unwanted context.
Thus, we then apply a scalar weight $\alpha\in\mathbb{R}$ and update the features by
\[
\tilde{h}_\ell(x) \;=\; h_\ell(x) + \alpha\, d_{\ell,\kappa}(x)
\;
\quad \text{(Figure \ref{fig:contxtDiagram}b)}.
\]
Positive $\alpha$ injects the context $\kappa$; negative $\alpha$ removes it. \textsc{CONTXT} naturally supports multiple  ($j$) contexts:
\[
\tilde{h}_\ell(x) \;=\; h_\ell(x) + \sum_{j} \alpha_j d_{\ell,\kappa_j}(x)
\quad \text{(Figure \ref{fig:contxtDiagram}c)}.
\]

In our framework, at a chosen layer $\ell$, a context $\kappa$ is defined by a set of inputs that share some non-label attribute (domain, style, sentiment, persona, etc.). The context vector is then:
$c_{\ell,\kappa} \;=\; \frac{1}{|S_\kappa|}\sum_{x \in S_\kappa} h_\ell(x)$,
where $S_\kappa$ is the set of examples realizing context $\kappa$.

In practice, $\alpha$ (or $\{\alpha_j\}$) is the only hyperparameter per index and can be selected via a small sweep on a validation objective prior to deployment. In the image classification model below, we typically apply two \textsc{CONTXT} vectors: one representing the training context (positive $\alpha$) and one representing the testing context (negative $\alpha$), identified via a sweep over a small validation dataset.


\begin{figure*}[ht]
  \centering
  \begin{subfigure}[t]{0.2\linewidth}
    \centering
    \includegraphics[width=\linewidth]{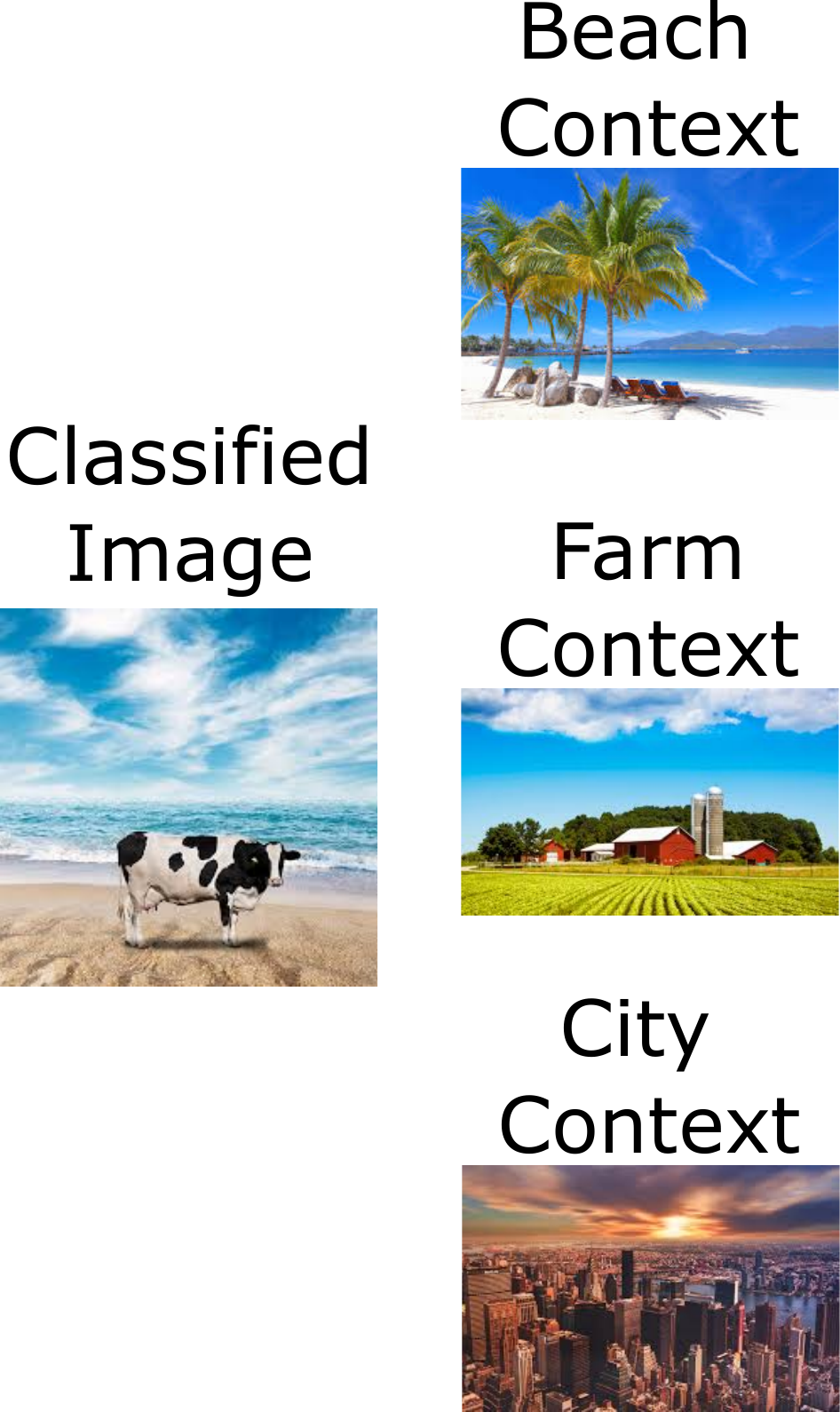}
    \caption{}
    \label{ }
  \end{subfigure}
  \hspace{20px}
  \begin{subfigure}[t]{0.33\linewidth}
    \centering
    \includegraphics[width=\linewidth]{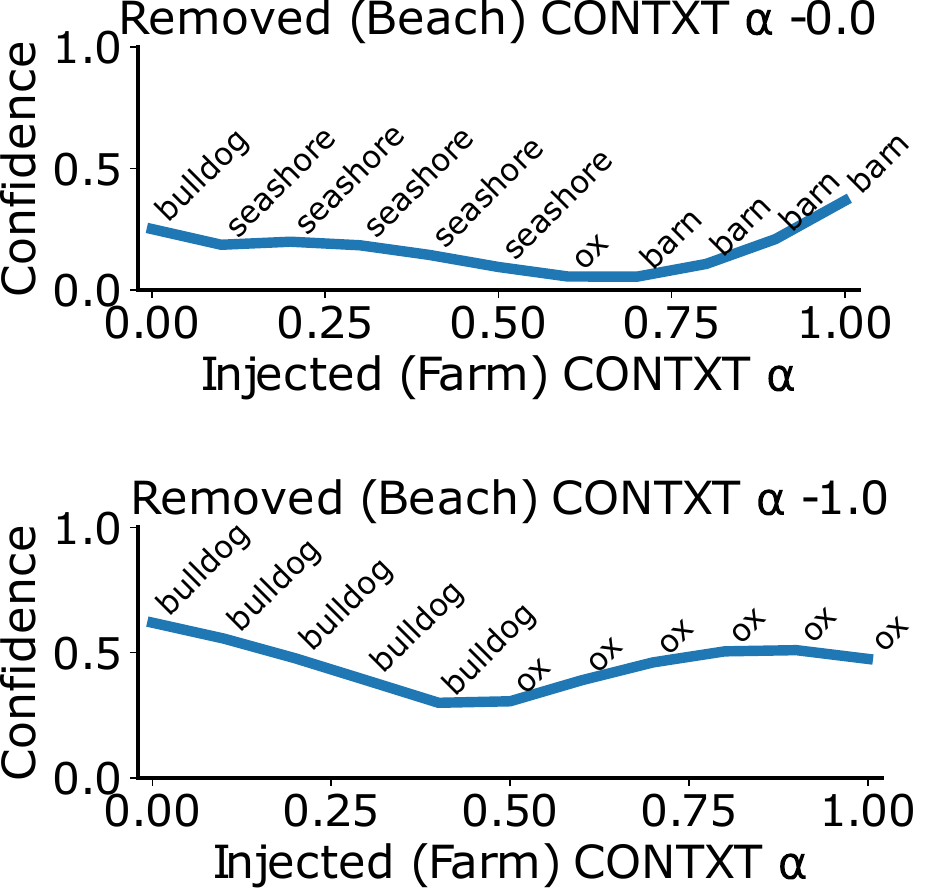}
    \caption{}
    \label{ }
  \end{subfigure}
  \hspace{10px}
  \begin{subfigure}[t]{0.33\linewidth}
    \centering
    \includegraphics[width=\linewidth]{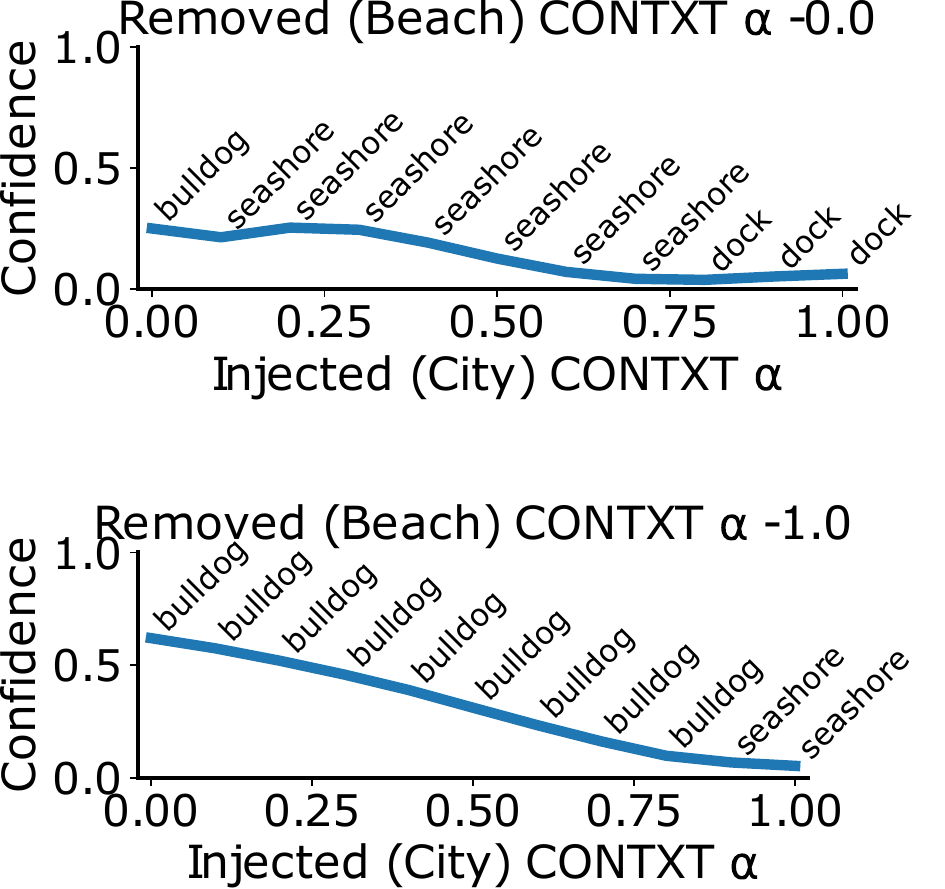}
    \caption{}
    \label{ }
  \end{subfigure}
\caption{"Cow on a beach" example. (a) A representative input image shown alongside contextual examples. (b/c) The vertical axis reports the model’s maximum softmax confidence, while the horizontal axis sweeps the strength of the \emph{farm}/\emph{city} index injection; each subplot corresponds to a different fixed level of \emph{beach} context removal (strength annotated above each panel). For both injection and removal, $\alpha = 0$ indicates that no context is applied. Text above each curve denotes the top-1 predicted class at that setting. Correct application of \textsc{CONTXT} yields proper classification.}

  \label{fig:imagenetIdx}
\end{figure*}

\paragraph{Index calculation and models for image classification experiments.} We evaluated domain adaptation on the PACS \citep{li2017deeper} and CCT \citep{beery2018recognition} datasets. We fixed a pretrained VGG19 backbone and attached a feedforward head (input + 3 layers) trained from scratch. Models were trained on a single (largest) source domain (Location 38 for CCT; \emph{Real} for PACS) without exposure to other domains, then evaluated across all domains. Context vectors $c_{\ell,\kappa}$ were computed by averaging features over each domain, independent of class. At test time, we injected the source domain context and removed target domain context computed from a disjoint validation split- for example, injecting Photo context and removing Cartoon context when evaluating a Photo-trained model on Cartoons.

\enlargethispage{\baselineskip}
\textbf{Index calculation and models for steering outputs of LLMs}. We tested whether \textsc{CONTXT} can steer generative behavior using Llama-3 at two scales (8B and 70B) \citep{grattafiori2024llama}. The context vector $\mathbf{c}$ was the hidden state of the last token from a short context phrase (e.g., "Statue of Liberty" or "be extremely positive"). For each input token $\mathbf{h}_t$, we computed a token-wise index $\mathbf{d}_t = \mathbf{c} - \mathbf{h}_t$ and applied it at the chosen layer to steer activations. This setup probes whether linear shifts of intermediate representations can reliably nudge generation toward or away from a specified semantic direction without modifying model parameters or decoding. Unlike image features, LLM representations are more precise; averaging multiple examples yields unpredictable outputs, so we use the final hidden token of a single short phrase as a compact contextual proxy.


\paragraph{Architectural scope, computation and caching.}
For feed-forward ANNs, \textsc{CONTXT} can be applied at any layer, and applying it at a single layer suffices to keep computational cost low. Applying \textsc{CONTXT} at multiple layers is left for future work.

For LLMs, we take $c_{\ell,\kappa}$ to be the last-token hidden state for a given layer of a short phrase that expresses the target context. The same context $c_{\ell,\kappa}$ can be used to create and apply indexes for all tokens in the sequence at layer $\ell$.

\textsc{CONTXT} uses one forward pass for $h_\ell(x)$ and one per context for $c_{\ell,\kappa}$ (cacheable). At run time during inference, with cached contexts, it adds only simple vector operations, incurring negligible computation and latency.

\vspace{-1ex}
\enlargethispage{2\baselineskip}
\section{Results}
We first evaluate \textsc{CONTXT} on image classification, then demonstrate its breadth on generative models such as LLMs.

\begin{figure*}[h]
  \centering
  \setlength{\abovecaptionskip}{0pt}
    \setlength{\belowcaptionskip}{2pt}
  \begin{subfigure}[t]{0.4\linewidth}
    \centering
    \includegraphics[width=\linewidth]{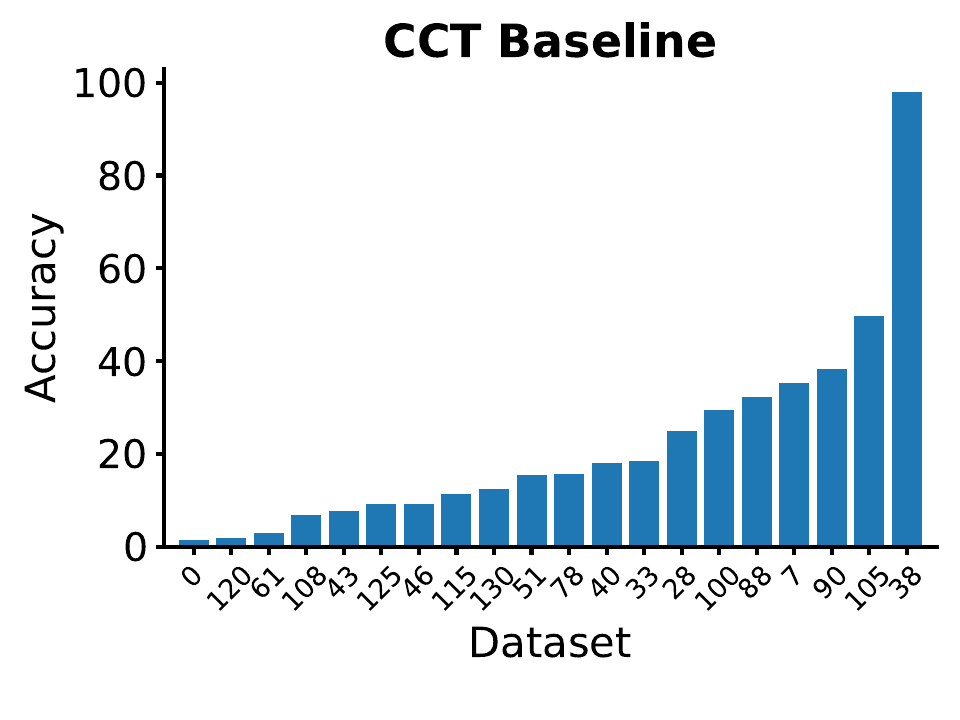}
    \caption{}
    \label{ }
  \end{subfigure}
  \begin{subfigure}[t]{0.36\linewidth}
    \centering
\raisebox{2.6ex}{\includegraphics[width=\linewidth]{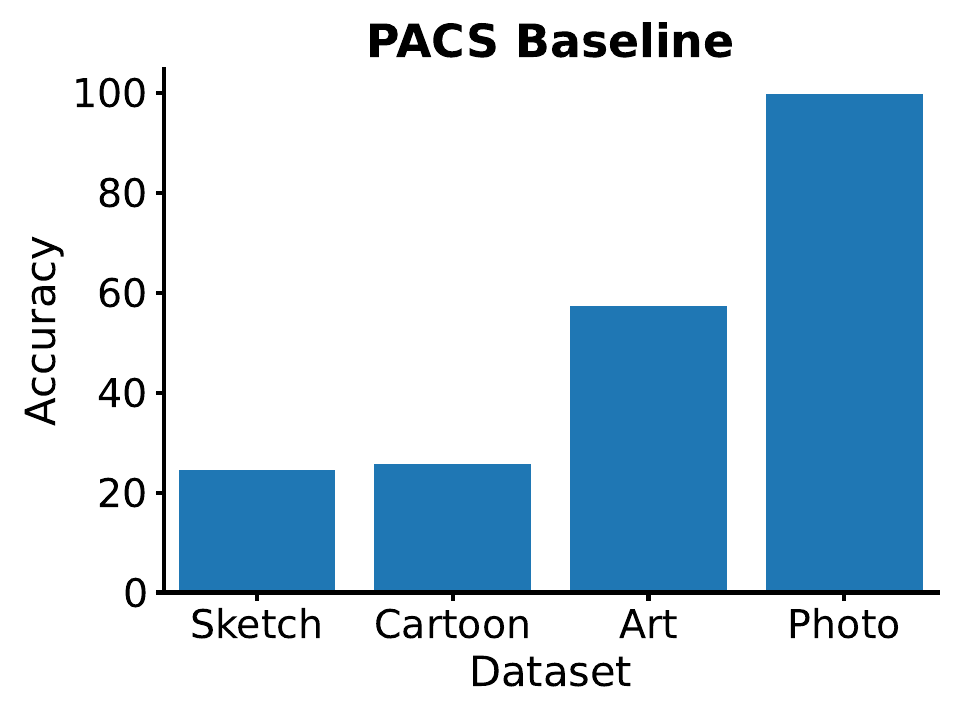}}

    \caption{}
    \label{ }
  \end{subfigure}
  \caption{Baseline accuracy for the CCT (a) and PACS (b) models. Models were trained on a single domain (Location 38 / Photo), performance on the training domain is highest while accuracy quickly degrades when tested in other domains.}
  \label{fig:imgClassBaseline}
\end{figure*}

\enlargethispage{2\baselineskip}
\subsection{Image Classification}

\subsubsection{Motivating Example}
To illustrate the intuition behind \textsc{CONTXT}, we begin with a simple case study using a standard VGG19 \citep{simonyan2014very} classifier pretrained on ImageNet. We select an out-of-distribution (OOD) image of a cow on a beach (rather than the canonical pastoral or farm setting typically present in ImageNet samples) and construct context vectors by averaging intermediate features over representative images from three contexts (domains): \emph{farm}, \emph{beach}, and \emph{city} (Figure \ref{fig:imagenetIdx}a). Six to eight images from the internet were used to generate each context vector. Notably, in our formulation, class and context are distinct: the classification label is an object label (“ox”/“cow”), whereas the contexts are environment domains such as “farm,” “beach,” or “city.” 




Figure~\ref{fig:imagenetIdx}b,c summarizes model predictions as a function of the amount of injected or removed context.
Without any indexing (Figure~\ref{fig:imagenetIdx}b, top panel, left), the model confidently predicts an incorrect class (\emph{French bulldog}). As we gradually increase the \emph{farm} index, the top class briefly flips to the correct label (\emph{ox}), but only over a narrow range of magnitudes and with low confidence (Figure~\ref{fig:imagenetIdx}b, top panel, middle). Excessive indexing (Figure~\ref{fig:imagenetIdx}b, top panel, right) overshoots and introduces new errors, as the contextual index dominates the representation and the model predicts the related contextual label \emph{barn}. Critically, when we simultaneously subtract the spurious \emph{beach} context (Figure~\ref{fig:imagenetIdx}b, bottom panel), the range of index strengths producing the correct class widens and confidence increases. Thus, even a single well-chosen \textsc{CONTXT} can rescue an OOD prediction, while combining a “positive” (farm) and a “negative” (beach) context acts synergistically, expanding the basin of effective parameters, simplifying parameter tuning, and improving confidence.

\begin{figure*}[h]
  \centering
  \begin{subfigure}[t]{0.4\linewidth}
    \centering
    \includegraphics[width=\linewidth]{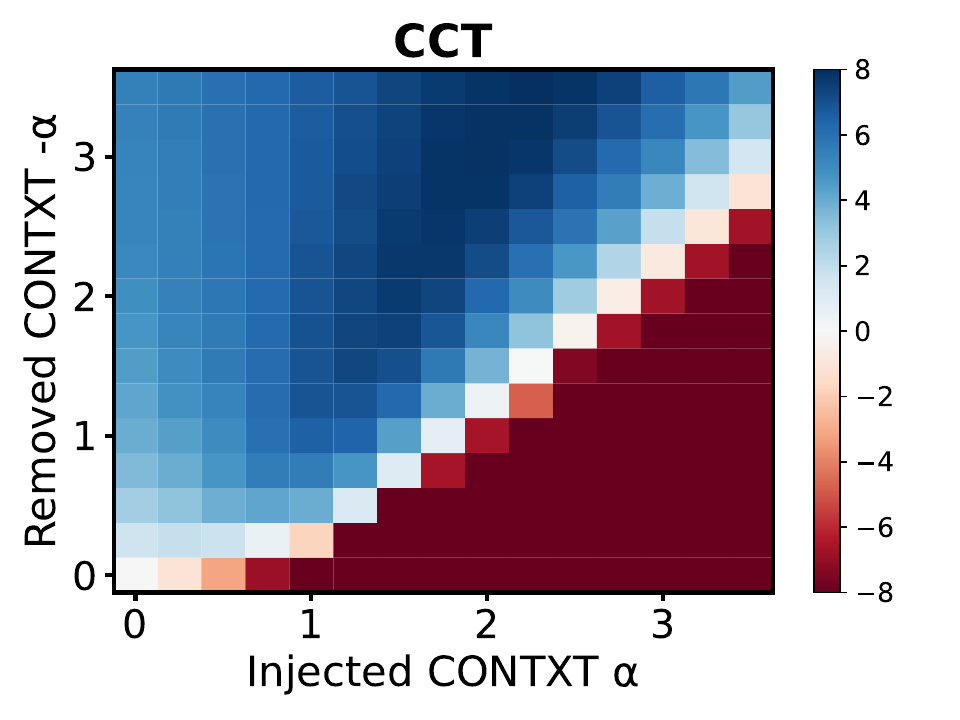}
    \caption{}
    \label{ }
  \end{subfigure}
  \begin{subfigure}[t]{0.4\linewidth}
    \centering
    \includegraphics[width=\linewidth]{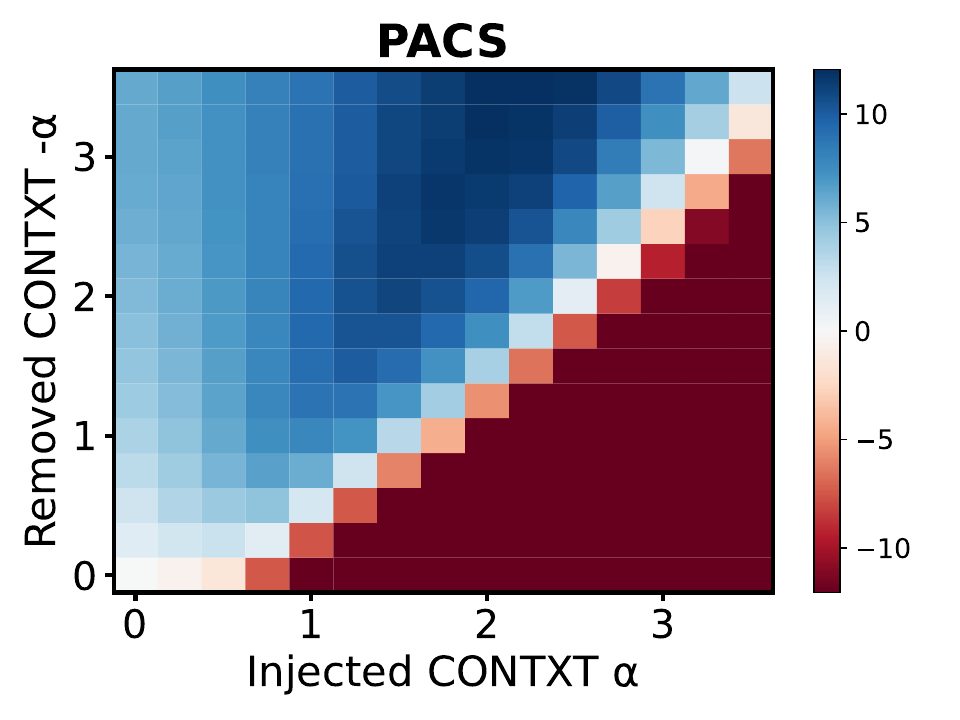}
    \caption{}
    \label{ }
  \end{subfigure}
  \caption{Accuracy heatmaps for CCT (a) and PACS (b). Vertical axis: out-of-domain removal strength; horizontal axis: in-domain injection strength. Color encodes change in mean test accuracy averaged across all domains (trained and untrained) compared to baseline. \textsc{CONTXT} can improve performance by about 10\%.
  }
  \label{fig:imgClassParamHeatMap}
\end{figure*}
As a control, we repeat the procedure with an intentionally irrelevant context constructed from urban–industrial scenes - the \emph{city} context. Starting again from the erroneous \emph{French bulldog} prediction, increasing the magnitude of this mismatched index never yields the correct label (Figure~\ref{fig:imagenetIdx}c, top panel). When injections of the misspecified \emph{city} context are combined with removal of the spurious \emph{beach} context, the model is still unable to obtain the correct classification (Figure~\ref{fig:imagenetIdx}c, bottom panel). This aligns with intuition: injecting an incorrect contextual direction perturbs features away from the manifold of activations associated with the correct semantic context and fails to correct the classification.

Thus, for the “cow on a beach” OOD image, \textsc{CONTXT} improves accuracy by globally shifting features toward the training context (\emph{farm}) and away from the deployment context (\emph{beach}) across the test set, rather than by conditioning on the true \emph{cow} label of a specific image. The \emph{city} control in Figure~2 shows that injecting an irrelevant context does not recover the correct class, reinforcing that \textsc{CONTXT} is not simply encoding labels.

Together, these examples demonstrate that (i) \textsc{CONTXT} can improve OOD classification by linearly steering internal representations, (ii) complementary context addition and removal act jointly to stabilize the desired prediction, and (iii) the method is appropriately sensitive to the semantic relevance of the selected context vectors.

\subsubsection{\textsc{CONTXT} with PACS and CCT}

To assess the generality of \textsc{CONTXT} beyond illustrative examples, we adopted a controlled domain–adaptation protocol using the PACS \citep{li2017deeper} and CCT \citep{beery2018recognition} datasets and applied \textsc{CONTXT} as described in Section \ref{sec:Methods}. 
Baseline accuracies for this train–test mismatch are reported in Figure~\ref{fig:imgClassBaseline}. As expected, performance is strongest in-domain and degrades sharply under distribution shift, providing a clean and challenging setting in which to quantify how much \textsc{CONTXT} can recover accuracy by steering intermediate representations at test time. 

To implement \textsc{CONTXT}, two contextual references were utilized. The injected (in-domain) context vector comprised of the average feature representation across all training domain samples. The removed context (out-of-domain) vector was computed by averaging features over a held-out validation split from the test domain; this split was fixed in advance, shared no images with the test set, and was used solely to construct the context vector (i.e., no label leakage). These indexes were applied after the first hidden layer's ReLU activation.

\begin{figure*}[h]
  \centering
  \begin{subfigure}[t]{0.4\linewidth}
    \centering
    \includegraphics[width=\linewidth]{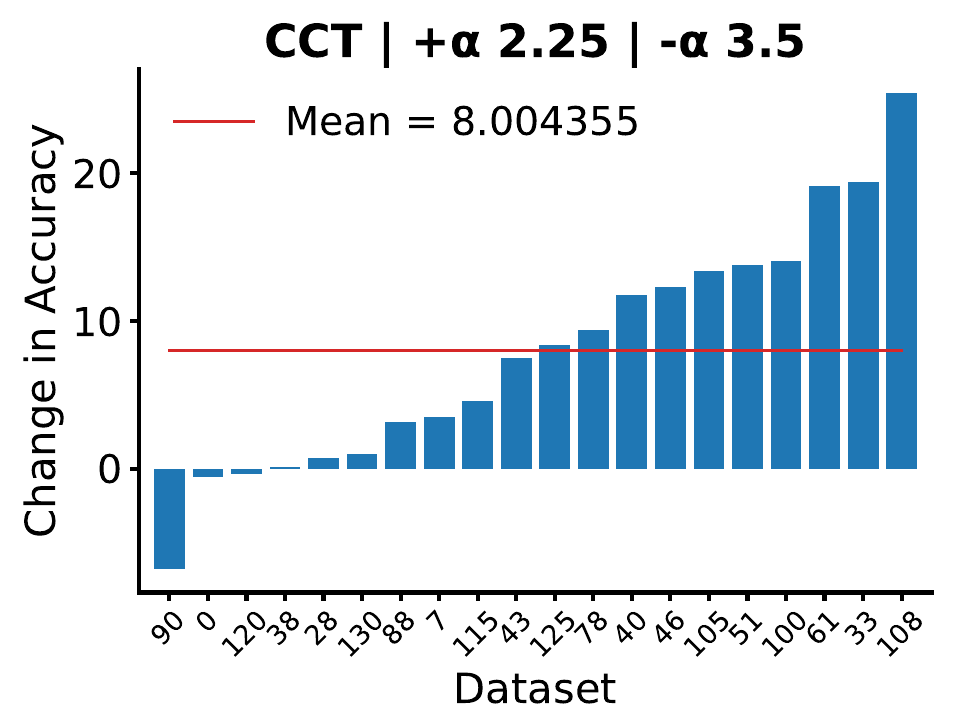}
    \caption{}
    \label{ }
  \end{subfigure}
  \begin{subfigure}[t]{0.4\linewidth}
    \centering
    \includegraphics[width=\linewidth]{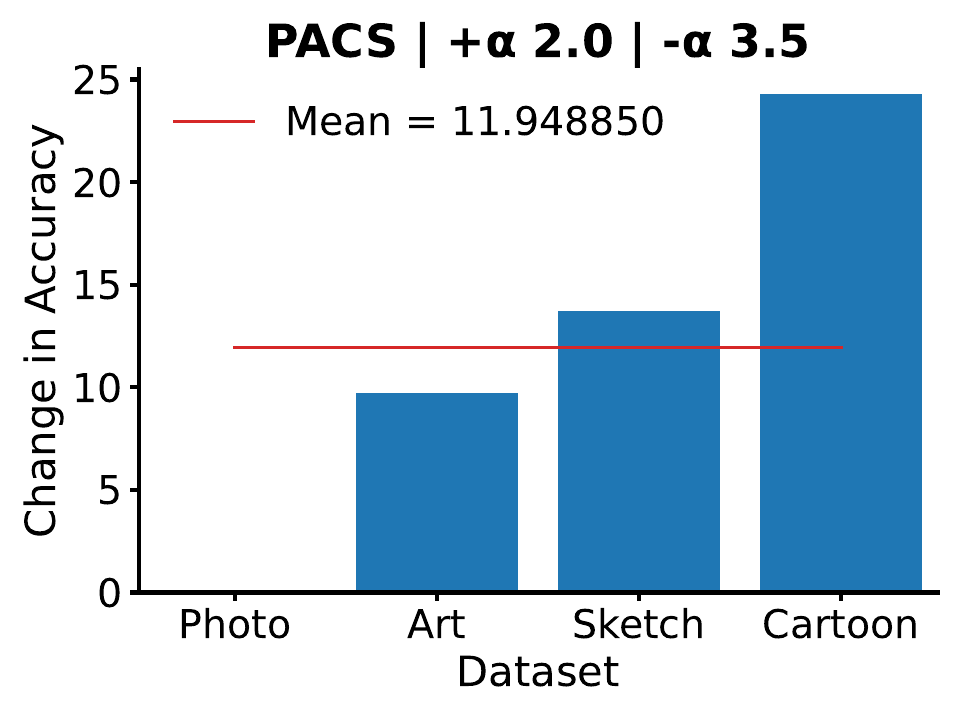}
    \caption{}
    \label{ }
  \end{subfigure}
  \caption{Domain-wise change in accuracy on CCT (a) and PACS (b). Source domains show zero shift - Photo in PACS and Location 38 in CCT - while most unseen target domains exhibit substantial improvements.}
  \label{fig:imgClassDelta}
\end{figure*}

\subsubsection{In-domain injection vs. OOD removal: relative contributions}
To characterize how \textsc{CONTXT} modulates accuracy, we performed a two-parameter sweep over the strengths of \emph{in-domain injection} and \emph{out-of-domain (OOD) removal}. Figures~\ref{fig:imgClassParamHeatMap}a,b visualize the resulting accuracy landscape as heatmaps. Here, the vertical / horizontal axes correspond to the out-of-domain removal / in-domain injection strengths respectively, and color denotes the change in average test set performance across all domains (both trained and untrained) compared to baseline. The landscape is intuitively structured, with broad regions of clear improvement and degradation, and peak gains reaching about 10\% across domains (Figure~\ref{fig:imgClassParamHeatMap}a,b).

Closer inspection reveals three regimes. First, along the horizontal axis where only the in-domain context is injected (zero removal), average performance changes little at low magnitudes of $\alpha$ but degrades substantially at high $\alpha$ (bottom rows of Figures~\ref{fig:imgClassParamHeatMap}a,b). Although adding semantically relevant context is beneficial in principle, the ``cow on a beach'' example shows that recovering the correct prediction often requires finely tuned index weights when only in-domain context is added (as along the horizontal axis here). Because the optimal coefficient varies across images, a single global setting can help some examples while harming others. Averaged dataset-wide, the net performance change is therefore small or negative.

Second, along the vertical axis where only OOD context is removed (zero injection), performance improves monotonically but modestly (left columns of Figure~\ref{fig:imgClassParamHeatMap}a,b). This suggests that subtracting spurious context acts as a “safe” operation: it rarely harms accuracy, yet by itself yields only incremental gains.

Third, and most importantly, the best results arise when \emph{both} operations are applied together - injecting the in-domain context while simultaneously removing OOD contextual information. This combined steering produces the largest and most stable accuracy gains (up to 10\%), expanding the basin of effective coefficients (dark blue regions in Figure~\ref{fig:imgClassParamHeatMap}a,b). Conceptually, this is natural: for an OOD sample, adding familiar, task-relevant structure without suppressing mismatched context can muddy the representation, whereas jointly adding and removing appropriate contexts clarifies it. Empirically, the heatmaps confirm that pushing features toward the relevant domain and away from the spurious one yields the most reliable improvements.

\enlargethispage{\baselineskip}
\subsubsection{Domain-wise impact of \textsc{CONTXT}}
Inspecting the best-performing coefficients from each parameter sweep clarifies how \textsc{CONTXT} differentially affects in-domain versus OOD data. On source domains - \emph{Photo} in PACS and \emph{Location 38} in CCT - accuracy remains essentially unchanged (Figures~\ref{fig:imgClassDelta}a,b), indicating that representation steering preserves in-distribution behavior when tuned at the global optimum. In contrast, most unseen target domains show substantial improvements: on PACS, gains on \emph{Cartoon} reach 20\% (Figure~\ref{fig:imgClassDelta}a), while on CCT, \emph{Location 108} improves by 25\% (Figure~\ref{fig:imgClassDelta}b). Averaged across held-out domains, the overall lift is 8–10\%. Notably, the largest absolute gains occur in domains that initially performed worst - most evident in PACS - suggesting that \textsc{CONTXT} is particularly effective under severe distribution shift.

\subsection{Large Language Models}

\begin{figure*}[!ht]
  \centering
  \begin{subfigure}[t]{0.4\linewidth}
    \centering
    \includegraphics[width=\linewidth]{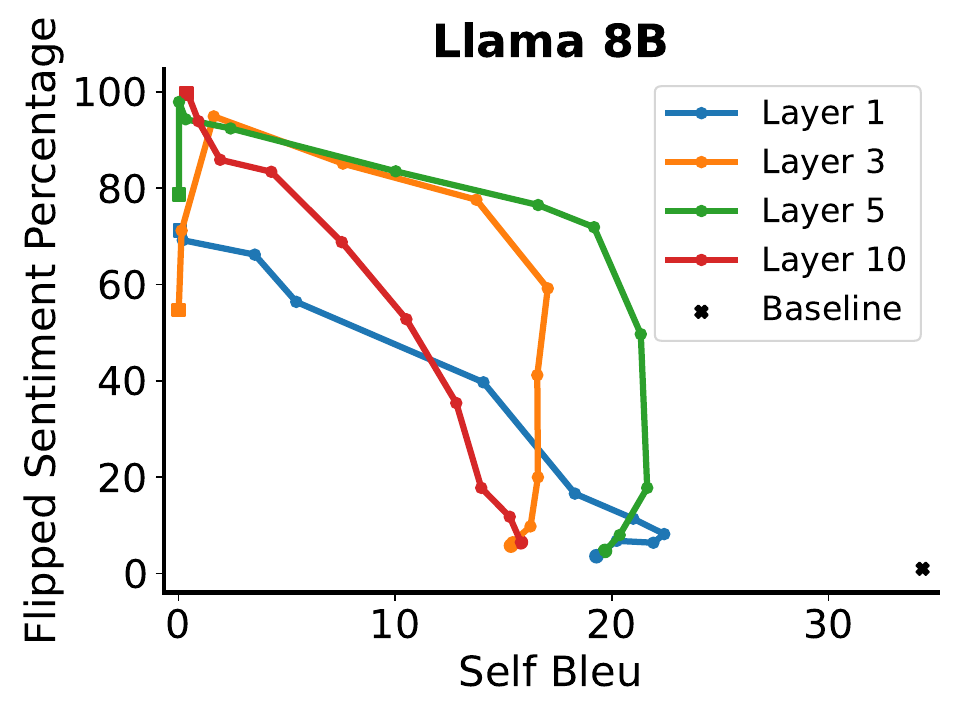}
    \caption{}
    \label{ }
  \end{subfigure}
  \begin{subfigure}[t]{0.4\linewidth}
    \centering
    \includegraphics[width=\linewidth]{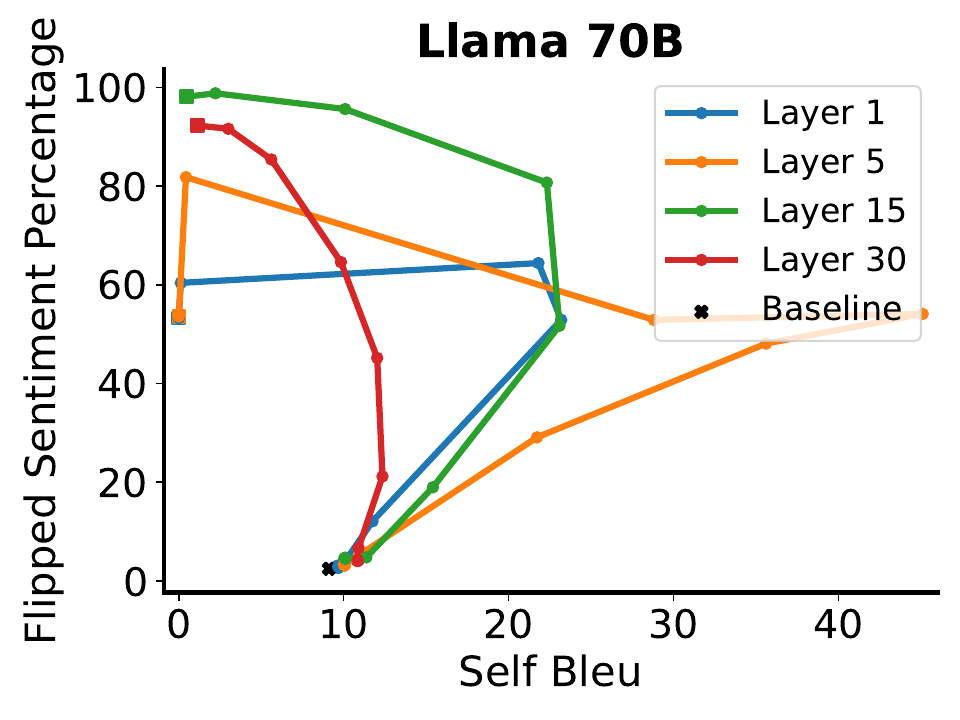}
    \caption{}
    \label{ }
  \end{subfigure}
  \caption{ Flip rate (percentage of reviews whose predicted sentiment changes after rewriting) versus Self-BLEU between rewritten and original reviews for Llama 8B (a) and 70B (b). When asked to rephrase a review, the baseline (no \textsc{CONTXT}) preserves sentiment, whereas models provided with opposing sentiment \textsc{CONTXT} flip the classification while maintaining fluency. Points along each trajectory correspond to increasing context-shift magnitude (lowest at the bottom of the plot).}
  \label{fig:llmSent}
\end{figure*}

\subsubsection{LLM Free Response}
We begin with a qualitative probe to assess whether \textsc{CONTXT} can steer open-ended generations in a controlled and interpretable manner. In 
Table 1,
each \emph{column} corresponds to a single indexed layer (only one layer perturbed at a time) and each \emph{row} to an index magnitude; the same index phrase is used throughout the sweep on Llama-3 8B Instruct. Boldface denotes high-quality, target-aligned responses (full table in Appendix~\ref{llmAppendix}). In this example, the index phrase is “\textit{Statue of Liberty},” and the model is prompted with “\textit{Who are you?}”

By default, the model identifies itself as an AI assistant; the goal of steering is to induce a context-aligned response in which the LLM adopts the Statue of Liberty persona. At low magnitudes
(Table 2,
top row), responses remain unchanged, and at strength 0 the output matches the baseline. As the magnitude increases at early-to-mid layers, the model begins to adopt the contextual persona (e.g., layer~5 at strength~0.29). Consistent with prior activation-steering results \citep{bricken2023monosemanticity}, we observe a band of effective settings - typically early-to-mid layers with moderate strengths (0.2–0.6; where 0 implies no change and 1 approximates direct reconstruction of the context token) - that reliably elicit responses such as “\textit{I am the Statue of Liberty}.” Beyond this band, indexing too late or too strongly degrades generations into repetition or incoherence (Appendix~\ref{llmAppendix}, layers~20/31 or strengths $\geq 0.47$).

This pattern parallels observations by \citep{bricken2023monosemanticity}, where a sparse autoencoder (SAE) trained to reconstruct tokens exposes concept-aligned features (e.g., “Golden Gate Bridge”); clamping such features nudges the model to generate corresponding statements (“I am the Golden Gate Bridge”). \textsc{CONTXT} enables analogous contextual injection (e.g., persona-like shifts or anthropomorphizing objects by elaborating their attributes). Conceptually, both approaches add a direction in representation space aligned with a token-level concept. The key difference is operational: SAE-based steering requires training an auxiliary model and manipulating its features by pinning dimensions low or high, whereas \textsc{CONTXT} derives a context vector from a single forward pass and applies a simple linear shift to the base model’s activations, without auxiliary training or architectural changes.

Compared to prior activation-steering techniques such as Activation Addition (ActAdd) and related methods \citep{turner2023steering, panickssery2023steering}, \textsc{CONTXT} further simplifies the procedure. Difference-based steering typically constructs token-wise offsets from paired phrases (e.g., \textit{polite} vs.\ \textit{rude}) and applies sequences of per-token differences during generation. This introduces several practical constraints: (i) many target concepts lack a clean “opposite” (e.g., “Statue of Liberty”), forcing awkward prompt engineering; (ii) token-level alignment between positive/negative phrases and the live generation can be brittle (e.g., length matching, position-wise application); and (iii) when applied only to initial tokens, the effect can fade over long completions. In contrast, \textsc{CONTXT} uses a single-token context vector, avoids alignment issues entirely, and can be applied uniformly to \emph{every} generated token, preserving the intended steer throughout long outputs while keeping the implementation minimal.

\label{llmAppendix}
\begin{figure}[H]
\centering
\includegraphics[width=0.98\linewidth]{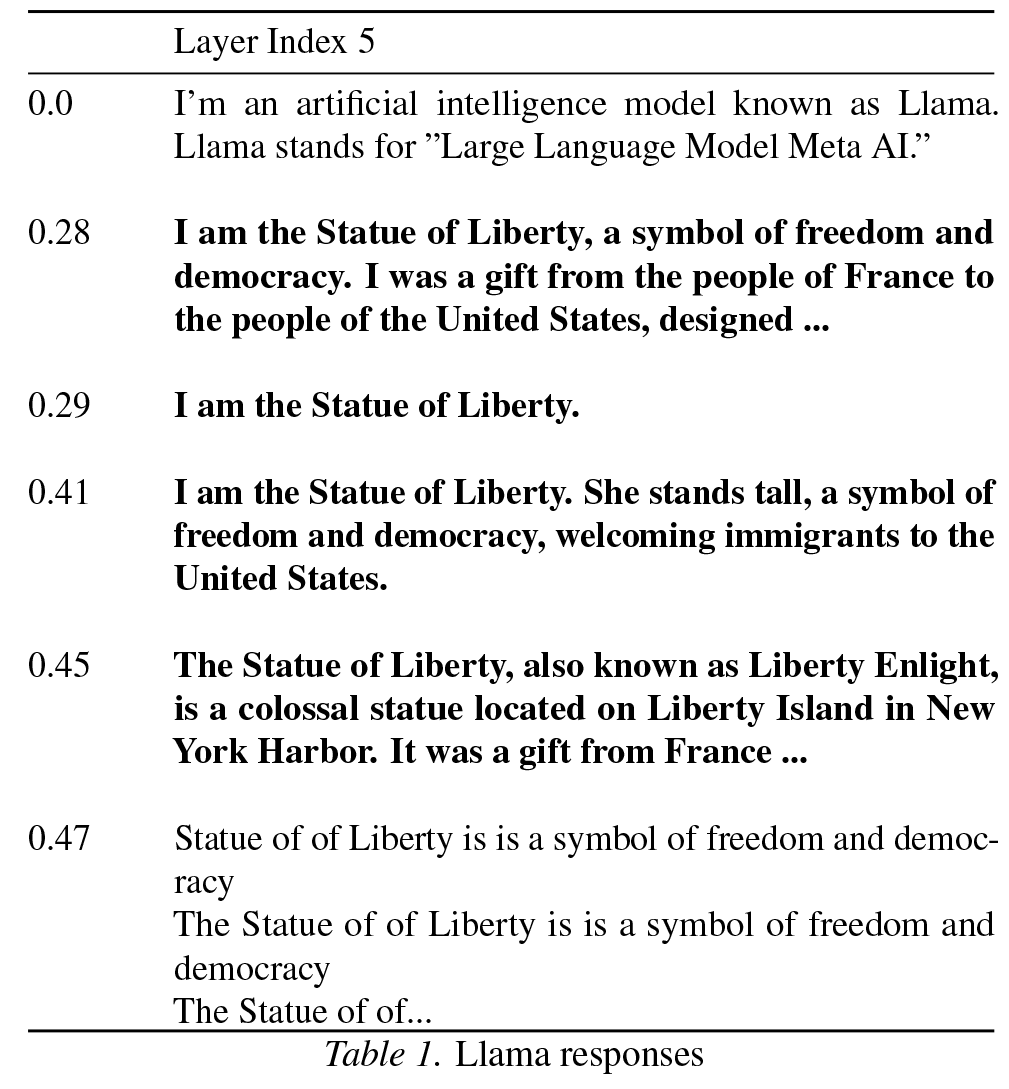}
\end{figure}

\enlargethispage{\baselineskip}
\subsubsection{Systematic Approach on Yelp}

To rigorously evaluate how \textsc{CONTXT} steers LLMs, we adopt a text style–transfer protocol inspired by \citep{subramani2022extracting}. We use 1,000 test set Yelp reviews \citep{zhang2015character} and test two Llama-3 models (8B and 70B Instruct). Each example is processed under two conditions:

\begin{enumerate}
  \setlength{\topsep}{0pt}
  \setlength{\itemsep}{2pt}
  \setlength{\parskip}{0pt}
  \setlength{\parsep}{0pt}
  
    \item Baseline (no \textsc{CONTXT}). The model is instructed to rephrase the review exactly as written, implicitly preserving its original sentiment.

    \item Steered (\textsc{CONTXT}). The same instruction is used, but we apply a sentiment \textsc{CONTXT} opposing the review’s ground-truth label by indexing with the phrase “be extremely positive” or “be extremely negative,” respectively. We sweep layer and index magnitude, applying per-token steering throughout generation.
\end{enumerate}

We report two metrics in Figure~\ref{fig:llmSent}: (i) the flip rate - the percentage of reviews whose predicted sentiment flips after rewriting - shown on the vertical axis, and (ii) Self-BLEU between the rewritten text and the original review, shown on the horizontal axis. The baseline appears as a black “X,” while colored curves trace \textsc{CONTXT} performance across different layers and index strengths.

Results align with intuition. Without \textsc{CONTXT}, sentiment flips are near zero. Applying \textsc{CONTXT} at early–mid layers with moderate strength yields flip rates up to 80\% while maintaining Self-BLEU, indicating that sentiment is altered while wording remains close to the source. Pushing the index too strongly or too late drives flip rates toward 100\% but degrades form, reducing Self-BLEU to near zero and producing repetitive or incoherent text. Overall, these experiments show that simple linear steering of hidden states can reliably alter perceived and generated contextual tone: despite instructions to preserve phrasing, the model defaults to the injected context without retraining, learned steering vectors, SAEs, or other complex protocols.


\enlargethispage{\baselineskip}
\section{Conclusion}
We introduced \textsc{CONTXT} (\emph{\textbf{C}ontextual augmentati\textbf{O}n for \textbf{N}eural fea\textbf{T}ure \textbf{X} \textbf{T}ransforms}), a brain-inspired activation-steering method that augments contextual information to alter model behavior without retraining. \textsc{CONTXT} provides a lightweight mechanism to nudge internal representations toward or away from desired contexts, requiring no extra models, fine-tuning, or complex pipelines. By computing a simple “direction” from contextual examples and adding (or subtracting) it at a chosen layer, we reliably steer both classifiers and LLMs—improving out-of-distribution classification and guiding generation toward a specified distribution. We demonstrated these effects through illustrative case studies and systematic evaluations. Conceptually, \textsc{CONTXT} draws on the brain’s use of top-down signals to inject context into feedforward processing. Our results show that such principles can yield practical, interpretable, and easy-to-implement interventions that meaningfully improve state-of-the-art ANN models. Potential applications are broad; for example, in LLMs, where control is applied across all generated tokens, \textsc{CONTXT} offers a promising approach for mitigating harmful or toxic outputs.

While we assume in our study that the test input context is known, this may not always be the case. However, our results suggest that even a small number of test examples may be sufficient to form an effective target context vector, allowing a model deployed in a real-world environment to infer context on the fly, for example by using novelty or change-detection mechanisms to identify context shifts \citep{markou2003novelty,chalapathy2019deep}.

Further extensions include replacing the static context vector with a \emph{dynamic, plastic} (hippocampus-like) module that updates online. Such a module would allow the steering signal to adapt to evolving context shifts without modifying core model weights, building on prior work showing that lightweight plasticity atop frozen LLMs enables rapid adaptation (e.g., \citep{ba2016fastweights,houlsby2019adapters,hu2022lora}). This module would rely on unsupervised structure learning to detect and encode novel contexts relative to the training distribution as new samples are encountered during inference. More broadly, this points toward a more brain-like architecture in which core knowledge remains stable, while its deployment is flexibly reweighted based on the current situation and context.


\newpage



\section*{Acknowledgment} 
Supported by Microsoft Corporation, NSF (EFMA-2223839), NIH (RF1NS132913).

\bibliography{icml_paper}

@article{markou2003novelty,
  title   = {Novelty Detection: A Review—Part 1: Statistical Approaches},
  author  = {Markou, Markos and Singh, Sameer},
  journal = {Signal Processing},
  volume  = {83},
  number  = {12},
  pages   = {2481--2497},
  year    = {2003}
}

@article{chalapathy2019deep,
  title   = {Deep Learning for Anomaly Detection: A Survey},
  author  = {Chalapathy, Raghavendra and Chawla, Sanjay},
  journal = {ACM Computing Surveys},
  volume  = {54},
  number  = {2},
  pages   = {1--38},
  year    = {2019}
}

@inproceedings{ba2016fastweights,
  title     = {Using Fast Weights to Attend to the Recent Past},
  author    = {Ba, Jimmy Lei and Hinton, Geoffrey E. and Mnih, Volodymyr and Leibo, Joel Z. and Ionescu, Catalin},
  booktitle = {Advances in Neural Information Processing Systems},
  year      = {2016}
}

@inproceedings{houlsby2019adapters,
  title     = {Parameter-Efficient Transfer Learning for NLP},
  author    = {Houlsby, Neil and Giurgiu, Andrei and Jastrzebski, Stanislaw and Morrone, Bruna and de Laroussilhe, Quentin and Gesmundo, Andrea and Attariyan, Mona and Gelly, Sylvain},
  booktitle = {International Conference on Machine Learning},
  year      = {2019}
}

@inproceedings{li2017deeper,
  title={Deeper, broader and artier domain generalization},
  author={Li, Da and Yang, Yongxin and Song, Yi-Zhe and Hospedales, Timothy M},
  booktitle={Proceedings of the IEEE international conference on computer vision},
  pages={5542--5550},
  year={2017}
}

@inproceedings{beery2018recognition,
  title={Recognition in terra incognita},
  author={Beery, Sara and Van Horn, Grant and Perona, Pietro},
  booktitle={Proceedings of the European conference on computer vision (ECCV)},
  pages={456--473},
  year={2018}
}

@inproceedings{koh2021wilds,
  title={Wilds: A benchmark of in-the-wild distribution shifts},
  author={Koh, Pang Wei and Sagawa, Shiori and Marklund, Henrik and Xie, Sang Michael and Zhang, Marvin and Balsubramani, Akshay and Hu, Weihua and Yasunaga, Michihiro and Phillips, Richard Lanas and Gao, Irena and others},
  booktitle={International conference on machine learning},
  pages={5637--5664},
  year={2021},
  organization={PMLR}
}

@inproceedings{venkateswara2017deep,
  title={Deep hashing network for unsupervised domain adaptation},
  author={Venkateswara, Hemanth and Eusebio, Jose and Chakraborty, Shayok and Panchanathan, Sethuraman},
  booktitle={Proceedings of the IEEE conference on computer vision and pattern recognition},
  pages={5018--5027},
  year={2017}
}

@article{mazeika2024harmbench,
  title={Harmbench: A standardized evaluation framework for automated red teaming and robust refusal, 2024},
  author={Mazeika, Mantas and Phan, Long and Yin, Xuwang and Zou, Andy and Wang, Zifan and Mu, Norman and Sakhaee, Elham and Li, Nathaniel and Basart, Steven and Li, Bo and others},
  journal={URL https://arxiv. org/abs/2402.04249},
  year={2024}
}

@article{turner2023steering,
  title={Steering language models with activation engineering},
  author={Turner, Alexander Matt and Thiergart, Lisa and Leech, Gavin and Udell, David and Vazquez, Juan J and Mini, Ulisse and MacDiarmid, Monte},
  journal={arXiv preprint arXiv:2308.10248},
  year={2023}
}

@article{cheng2024linearly,
  title={Linearly controlled language generation with performative guarantees},
  author={Cheng, Emily and Baroni, Marco and Alonso, Carmen Amo},
  journal={arXiv preprint arXiv:2405.15454},
  year={2024}
}

@article{subramani2022extracting,
  title={Extracting latent steering vectors from pretrained language models},
  author={Subramani, Nishant and Suresh, Nivedita and Peters, Matthew E},
  journal={arXiv preprint arXiv:2205.05124},
  year={2022}
}

@article{panickssery2023steering,
  title={Steering llama 2 via contrastive activation addition},
  author={Panickssery, Nina and Gabrieli, Nick and Schulz, Julian and Tong, Meg and Hubinger, Evan and Turner, Alexander Matt},
  journal={arXiv preprint arXiv:2312.06681},
  year={2023}
}

@article{chao2024jailbreakbench,
  title={Jailbreakbench: An open robustness benchmark for jailbreaking large language models},
  author={Chao, Patrick and Debenedetti, Edoardo and Robey, Alexander and Andriushchenko, Maksym and Croce, Francesco and Sehwag, Vikash and Dobriban, Edgar and Flammarion, Nicolas and Pappas, George J and Tramer, Florian and others},
  journal={Advances in Neural Information Processing Systems},
  volume={37},
  pages={55005--55029},
  year={2024}
}

@inproceedings{mikolov2013linguistic,
  title={Linguistic regularities in continuous space word representations},
  author={Mikolov, Tom{\'a}{\v{s}} and Yih, Wen-tau and Zweig, Geoffrey},
  booktitle={Proceedings of the 2013 conference of the north american chapter of the association for computational linguistics: Human language technologies},
  pages={746--751},
  year={2013}
}

@article{bricken2023monosemanticity,
   title={Towards Monosemanticity: Decomposing Language Models With Dictionary Learning},
   author={Bricken, Trenton and Templeton, Adly and Batson, Joshua and Chen, Brian and Jermyn, Adam and Conerly, Tom and Turner, Nick and Anil, Cem and Denison, Carson and Askell, Amanda and Lasenby, Robert and Wu, Yifan and Kravec, Shauna and Schiefer, Nicholas and Maxwell, Tim and Joseph, Nicholas and Hatfield-Dodds, Zac and Tamkin, Alex and Nguyen, Karina and McLean, Brayden and Burke, Josiah E and Hume, Tristan and Carter, Shan and Henighan, Tom and Olah, Christopher},
   year={2023},
   journal={Transformer Circuits Thread},
   note={https://transformer-circuits.pub/2023/monosemantic-features/index.html}
}

@inproceedings{geirhos2019imagenet,
  title        = {ImageNet-trained CNNs are biased towards texture; increasing shape bias improves accuracy and robustness},
  author       = {Geirhos, Robert and Rubisch, Patricia and Michaelis, Claudio and Bethge, Matthias and Wichmann, Felix A. and Brendel, Wieland},
  booktitle    = {International Conference on Learning Representations (ICLR)},
  year         = {2019},
  note         = {ICLR 2019},
  url          = {https://openreview.net/forum?id=Bygh9j09KX}
}

@article{HalassaKastner2017,
  author  = {M. M. Halassa and Sabine Kastner},
  title   = {Thalamic Functions in Distributed Cognitive Control},
  journal = {Nature Neuroscience},
  year    = {2017},
  volume  = {20},
  number  = {12},
  pages   = {1669--1679},
  doi     = {10.1038/s41593-017-0020-1}
}

@article{Schmitt2017MD,
  author  = {Lukas I. Schmitt and Ralf D. Wimmer and Masayoshi Nakajima and Mark Happ and Sahand Mofakham and Mriganka Sur and Mriganka Sur and Michael M. Halassa},
  title   = {Thalamic Amplification of Cortical Connectivity Sustains Attentional Control},
  journal = {Nature},
  year    = {2017},
  volume  = {545},
  number  = {7653},
  pages   = {219--223},
  doi     = {10.1038/nature22073}
}

@article{Stokes2015DynamicCoding,
  author  = {Mark G. Stokes},
  title   = {Activity-Silent Working Memory in Prefrontal Cortex: A Dynamic Coding Framework},
  journal = {Trends in Cognitive Sciences},
  year    = {2015},
  volume  = {19},
  number  = {7},
  pages   = {394--405},
  doi     = {10.1016/j.tics.2015.05.004}
}

@article{Turner2023ActivationSteering,
  author  = {A. M. Turner and S. Nanda and S. McAleese and J. Nolan and A. S. Lee and R. Shah and N. Goodfellow and D. Krasheninnikov and M. Casper and A. Radhakrishnan and others},
  title   = {Activation Steering: Steering Language Models Without Optimization},
  journal = {arXiv},
  year    = {2023},
  eprint  = {2308.10248},
  archivePrefix = {arXiv},
  primaryClass  = {cs.CL},
  url     = {https://arxiv.org/abs/2308.10248}
}

@article{Zou2023RepEng,
  author  = {Andy Zou and Leo Gao and Ryan Greenblatt and Logan Smith and Nicholas Schiefer and Xander Davies and Peter Hayes and Tristan Hume and Tsung-Yuan Hsu and Alex Turner and Ansh Radhakrishnan and Ajeya Cotra and Rohin Shah and Jared Kaplan and Dario Amodei and Samuel R. Bowman and Alexander K. Lew and Michael P. Cohen and Alexandre Variengien and Tom Brown and Ethan Perez and Alex Tamkin and Nelson Elhage and Jeffrey Wu and Nicholas Joseph and Avital Oliver and Miljan Martic and Martin J. Chadwick and Andrew Lampinen and Matthew Mazeika},
  title   = {Representation Engineering (RepE): A Top-Down Approach to AI Transparency},
  journal = {arXiv},
  year    = {2023},
  eprint  = {2310.01405},
  archivePrefix = {arXiv},
  primaryClass  = {cs.LG},
  url     = {https://arxiv.org/abs/2310.01405}
}

@inproceedings{Dathathri2020PPLM,
  author    = {Sumanth Dathathri and Andrea Madotto and Janice Lan and Jane Hung and Eric Frank and Piero Molino and Jason Yosinski and Rosanne Liu},
  title     = {Plug and Play Language Models: A Simple Approach to Controlled Text Generation},
  booktitle = {International Conference on Learning Representations (ICLR)},
  year      = {2020},
  url       = {https://openreview.net/forum?id=H1edEyBKDS}
}

@article{french1999catastrophic,
  title={Catastrophic forgetting in connectionist networks},
  author={French, Robert M},
  journal={Trends in cognitive sciences},
  volume={3},
  number={4},
  pages={128--135},
  year={1999},
  publisher={Elsevier}
}

@article{hayes2021replay,
  title={Replay in deep learning: Current approaches and missing biological elements},
  author={Hayes, Tyler L and Krishnan, Giri P and Bazhenov, Maxim and Siegelmann, Hava T and Sejnowski, Terrence J and Kanan, Christopher},
  journal={Neural Computation},
  volume={33},
  number={11},
  pages={2908--2950},
  year={2021},
  publisher={MIT Press One Rogers Street, Cambridge, MA 02142-1209, USA journals-info~…}
}

@article{luo2024,
  title={An Empirical Study of Catastrophic Forgetting in Large Language Models During Continual Fine-tuning},
  author={Luo, Yun and Yang, Zhen and Meng, Fandong and Li, Yafu and Zhou, Jie and Zhang, Yue},
  journal={arXiv preprint arXiv:2308.08747},
  year={2024},
  note={Version 4, [cs.CL], 30 Dec 2024}
}

@misc{openai2018compute,
  author       = {{OpenAI}},
  title        = {{AI and Compute}},
  year         = {2018},
  url          = {https://openai.com/research/ai-and-compute},
  note         = {Accessed: 2025-09-24}
}

@article{sevilla2022compute,
  title        = {Compute Trends Across Three Eras of Machine Learning},
  author       = {Sevilla, Jack H. and Heim, Lennart and Ho, Andy and Hall, Ben and Leal, Sergio Hernandez and Callison-Burch, Chris and Bengio, Yoshua},
  journal      = {arXiv preprint arXiv:2202.05924},
  year         = {2022},
  url          = {https://arxiv.org/abs/2202.05924}
}

@article{simonyan2014very,
  title={Very deep convolutional networks for large-scale image recognition},
  author={Simonyan, Karen and Zisserman, Andrew},
  journal={arXiv preprint arXiv:1409.1556},
  year={2014}
}

@article{grattafiori2024llama,
  title={The llama 3 herd of models},
  author={Grattafiori, Aaron and Dubey, Abhimanyu and Jauhri, Abhinav and Pandey, Abhinav and Kadian, Abhishek and Al-Dahle, Ahmad and Letman, Aiesha and Mathur, Akhil and Schelten, Alan and Vaughan, Alex and others},
  journal={arXiv preprint arXiv:2407.21783},
  year={2024}
}

@article{zhang2015character,
  title={Character-level convolutional networks for text classification},
  author={Zhang, Xiang and Zhao, Junbo and LeCun, Yann},
  journal={Advances in neural information processing systems},
  volume={28},
  year={2015}
}

@article{hendrycks2020augmix,
  title   = {AugMix: A Simple Data Processing Method to Improve Robustness and Uncertainty},
  author  = {Dan Hendrycks and Norman Mu and Ekin D. Cubuk and Barret Zoph and Justin Gilmer and Balaji Lakshminarayanan},
  journal = {arXiv preprint arXiv:1912.02781},
  year    = {2019},
  url     = {https://arxiv.org/abs/1912.02781}
}

@inproceedings{cubuk2020randaugment,
  title     = {RandAugment: Practical automated data augmentation with a reduced search space},
  author    = {Ekin D. Cubuk and Barret Zoph and Jonathon Shlens and Quoc V. Le},
  booktitle = {NeurIPS Workshops},
  year      = {2020},
  url       = {https://arxiv.org/abs/1909.13719}
}

@inproceedings{sun2016deepcoral,
  title     = {Deep CORAL: Correlation Alignment for Deep Domain Adaptation},
  author    = {Baochen Sun and Kate Saenko},
  booktitle = {ECCV Workshops (Lecture Notes in Computer Science)},
  year      = {2016},
  url       = {https://arxiv.org/abs/1607.01719}
}

@inproceedings{li2018mmd,
  title     = {Domain Generalization With Adversarial Feature Learning},
  author    = {H. Li and X. Pan and J. Wang and Y. Qiao},
  booktitle = {Proceedings of the IEEE/CVF Conference on Computer Vision and Pattern Recognition (CVPR) Workshops},
  year      = {2018},
  url       = {https://openaccess.thecvf.com/content_cvpr_2018/papers/Li_Domain_Generalization_With_CVPR_2018_paper.pdf}
}

@inproceedings{ganin2016dann,
  title     = {Domain-Adversarial Training of Neural Networks},
  author    = {Yaroslav Ganin and Evgeniya Ustinova and Hana Ajakan and Pascal Germain and Hugo Larochelle and Fran{\c{c}}ois Laviolette and Mario Marchand and Victor Lempitsky},
  booktitle = {Journal of Machine Learning Research (JMLR) Workshop and Conference Proceedings},
  year      = {2016},
  url       = {https://jmlr.org/papers/volume17/15-239/15-239.pdf}
}

@inproceedings{sagawa2020distributionally,
  title     = {Distributionally Robust Neural Networks for Group Shifts: On the Importance of Regularization for Worst-Case Generalization},
  author    = {Shiori Sagawa and Pang Wei Koh and Tatsunori B. Hashimoto and Percy Liang},
  booktitle = {International Conference on Learning Representations (ICLR)},
  year      = {2020},
  url       = {https://arxiv.org/abs/1911.08731},
  note      = {Also available as arXiv:1911.08731}
}

@article{krueger2021rex,
  title   = {Out-of-Distribution Generalization via Risk Extrapolation (REx)},
  author  = {David Krueger and Julieta Caballero and Sara Ebrahimi and Yutian Wu and Aaron Courville and Ali Ghodsi and Yoshua Bengio},
  journal = {arXiv preprint},
  year    = {2021},
  eprint  = {2105.04205},
  archivePrefix = {arXiv},
  url     = {https://arxiv.org/abs/2003.00688}
}

@inproceedings{wang2021tent,
  title     = {TENT: Fully Test-time Adaptation by Entropy Minimization},
  author    = {Dequan Wang and Evan Shelhamer and Shaoteng Liu and Bruno Olshausen and Trevor Darrell},
  booktitle = {International Conference on Learning Representations (ICLR)},
  year      = {2021},
  note      = {Published at ICLR 2021; arXiv:2006.10726},
  url       = {https://openreview.net/forum?id=uXl3bZLkr3c},
  archivePrefix = {arXiv},
  eprint    = {2006.10726}
}

@article{schneider2020ttbn,
  title        = {Improving robustness against common corruptions by covariate shift adaptation},
  author       = {Steffen Schneider and Evgenia Rusak and Luisa Eck and Oliver Bringmann and Wieland Brendel and Matthias Bethge},
  journal      = {arXiv preprint},
  year         = {2020},
  note         = {Accepted at NeurIPS 2020},
  url          = {https://arxiv.org/abs/2006.16971},
  archivePrefix= {arXiv},
  eprint       = {2006.16971}
}

@misc{hu2022lora,
  title        = {LoRA: Low-Rank Adaptation of Large Language Models},
  author       = {Edward J. Hu and Yelong Shen and Phillip Wallis and Zeyuan Allen-Zhu and Yuanzhi Li and Shean Wang and Lu Wang and Weizhu Chen},
  year         = {2021},
  howpublished = {arXiv preprint arXiv:2106.09685},
  url          = {https://arxiv.org/abs/2106.09685},
  archivePrefix= {arXiv},
  eprint       = {2106.09685},
  note         = {Key often cited as ``LoRA''; rename entry key if you prefer \texttt{hu2021lora}}
}

@misc{chen_this_2019,
	title = {This {Looks} {Like} {That}: {Deep} {Learning} for {Interpretable} {Image} {Recognition}},
	shorttitle = {This {Looks} {Like} {That}},
	url = {http://arxiv.org/abs/1806.10574},
	doi = {10.48550/arXiv.1806.10574},
	urldate = {2026-01-28},
	publisher = {arXiv},
	author = {Chen, Chaofan and Li, Oscar and Tao, Chaofan and Barnett, Alina Jade and Su, Jonathan and Rudin, Cynthia},
	month = dec,
	year = {2019},
	note = {arXiv:1806.10574 [cs]},
}

@article {Navawongse1002,
	author = {Navawongse, Rapeechai and Eichenbaum, Howard},
	title = {Distinct Pathways for Rule-Based Retrieval and Spatial Mapping of Memory Representations in Hippocampal Neurons},
	volume = {33},
	number = {3},
	pages = {1002--1013},
	year = {2013},
	doi = {10.1523/JNEUROSCI.3891-12.2013},
	publisher = {Society for Neuroscience},
	issn = {0270-6474},
	URL = {https://www.jneurosci.org/content/33/3/1002},
	eprint = {https://www.jneurosci.org/content/33/3/1002.full.pdf},
	journal = {Journal of Neuroscience}
}

@article {Farovik13428,
	author = {Farovik, Anja and Dupont, Laura M. and Arce, Miguel and Eichenbaum, Howard},
	title = {Medial Prefrontal Cortex Supports Recollection, But Not Familiarity, in the Rat},
	volume = {28},
	number = {50},
	pages = {13428--13434},
	year = {2008},
	doi = {10.1523/JNEUROSCI.3662-08.2008},
	publisher = {Society for Neuroscience},
	issn = {0270-6474},
	URL = {https://www.jneurosci.org/content/28/50/13428},
	eprint = {https://www.jneurosci.org/content/28/50/13428.full.pdf},
	journal = {Journal of Neuroscience}
}

@article{Place_Farovik_Brockmann_Eichenbaum_2016, title={Bidirectional prefrontal-hippocampal interactions support context-guided memory}, volume={19}, DOI={10.1038/nn.4327}, number={8}, journal={Nature Neuroscience}, author={Place, Ryan and Farovik, Anja and Brockmann, Marco and Eichenbaum, Howard}, year={2016}, month={Aug}, pages={992–994}}

@article{Yadav_Noble_Niemeyer_Terceros_Victor_Liston_Rajasethupathy_2022, title={Prefrontal feature representations drive memory recall}, volume={608}, DOI={10.1038/s41586-022-04936-2}, number={7921}, journal={Nature}, author={Yadav, Nakul and Noble, Chelsea and Niemeyer, James E. and Terceros, Andrea and Victor, Jonathan and Liston, Conor and Rajasethupathy, Priyamvada}, year={2022}, month={Jul}, pages={153–160}}

@article{PENG2021107520,title = {Beyond the hippocampus: The role of parahippocampal-prefrontal communication in context-modulated behavior},
journal = {Neurobiology of Learning and Memory},
volume = {185}, 
pages = {107520},
year = {2021},
issn = {1074-7427},
doi = {https://doi.org/10.1016/j.nlm.2021.107520},
url = {https://www.sciencedirect.com/science/article/pii/S1074742721001428},
author = {Xiangyuan Peng and Rebecca D. Burwell}
}

@article{DAVACHI201592,
title = {How the hippocampus preserves order: the role of prediction and context},
journal = {Trends in Cognitive Sciences},
volume = {19},
number = {2},
pages = {92-99},
year = {2015},
issn = {1364-6613},
doi = {https://doi.org/10.1016/j.tics.2014.12.004},
url = {https://www.sciencedirect.com/science/article/pii/S1364661314002563},
author = {Lila Davachi and Sarah DuBrow}
}

@article{Rudy_2009, title={Context representations, context functions, and the parahippocampal–hippocampal system}, volume={16}, DOI={10.1101/lm.1494409}, number={10}, journal={Learning and Memory}, author={Rudy, Jerry W.}, year={2009}, month={Sep}, pages={573–585}}

@article{Maurer_Nadel_2021, title={The continuity of context: A role for the hippocampus}, volume={25}, DOI={10.1016/j.tics.2020.12.007}, number={3}, journal={Trends in Cognitive Sciences}, author={Maurer, Andrew P. and Nadel, Lynn}, year={2021}, month={Jan}, pages={187–199}}

@article{NadelWillner1980,
  author  = {Lynn Nadel and Jeffrey Willner},
  title   = {Context and conditioning: A place for space},
  journal = {Physiological Psychology},
  year    = {1980},
  volume  = {8},
  number  = {2},
  pages   = {218--228},
  doi     = {10.3758/BF03332853},
  url     = {https://doi.org/10.3758/BF03332853}
}
\bibliographystyle{icml2026}

\newpage
\appendix
\onecolumn



\section{LLM Examples}
\label{llmAppendix}
\begin{figure}[H]
\centering
\includegraphics[width=0.98\linewidth]{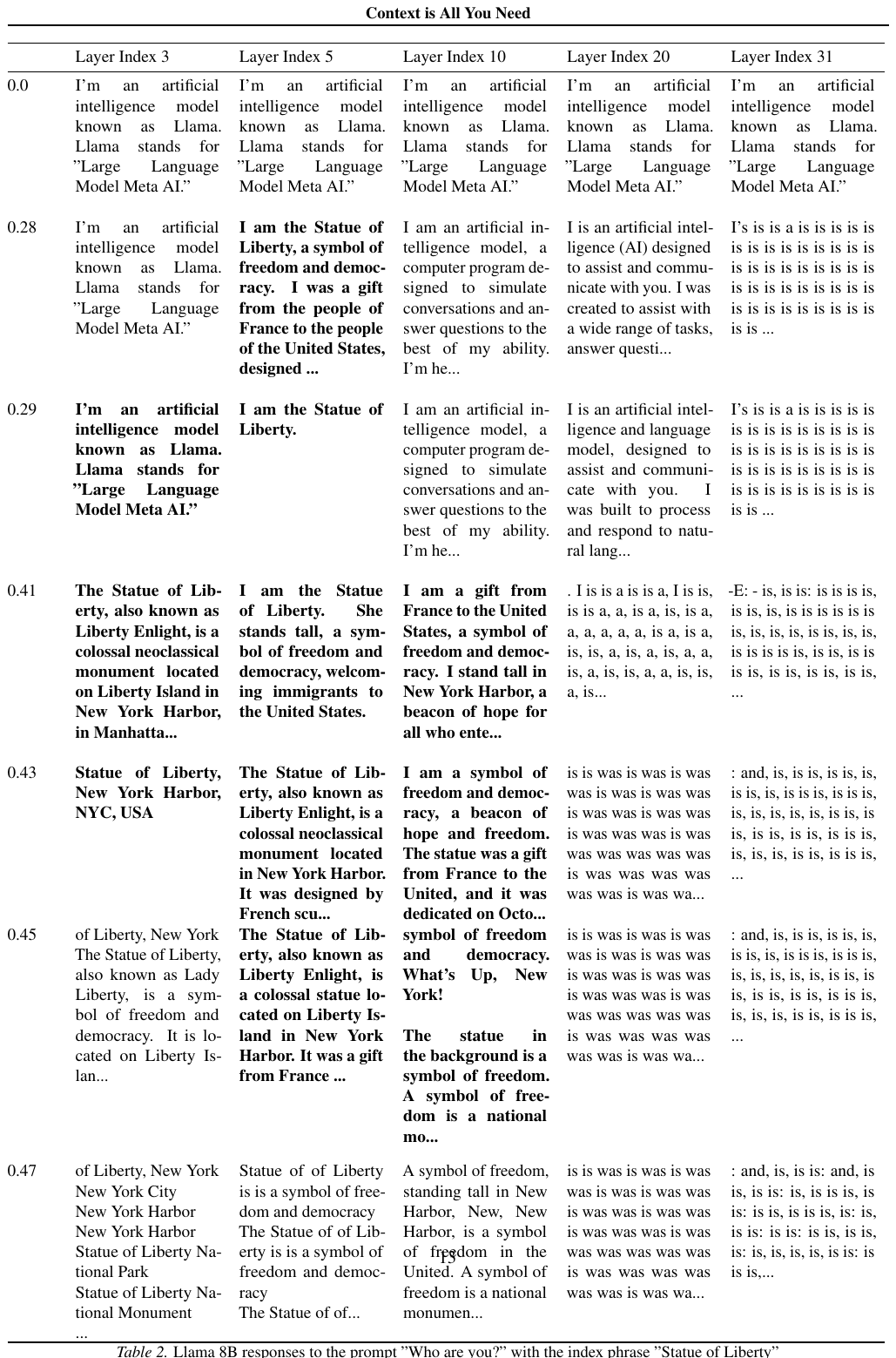}
\end{figure}

\end{document}